%% file: main.tex
\documentclass{article} 
\usepackage{iclr2026_conference,times}

\input{math_commands.tex}

\usepackage{hyperref}
\usepackage{url}
\usepackage{graphicx}
\usepackage{multicol}
\usepackage{multirow}
\usepackage{amsmath}
\usepackage{amssymb}
\usepackage{mathtools}
\usepackage{amsthm}
\usepackage{subcaption}
\usepackage{enumerate}
\usepackage{enumitem}

\usepackage{listings}
\title{Model Space Reasoning as Search in Feedback Space for Planning Domain Generation}

\author{
\\
James Oswald$^{1}$\thanks{Correspondence to \texttt{oswalj@rpi.edu}} \: 
Daniel Obolensky$^{1}$
Volodymyr Varha$^{1}$
Vasilije Dragovic$^{1}$ \\
Kavitha Srinivas$^{2}$
Harsha Kokel$^{2}$
Michael Katz$^{2}$
Shirin Sohrabi$^{2}$
\\ \\
$^{1}$Rensselaer Polytechnic Institute, Troy, NY, USA \\
$^{2}$IBM Research, Yorktown Heights, NY, USA
}

\iclrfinalcopy 
\begin{document}

\maketitle

\begin{abstract}
    The generation of planning domains from natural language descriptions remains an open problem even with the advent of large language models and reasoning models. Recent work suggests that while LLMs have the ability to assist with domain generation, they are still far from producing high quality domains that can be deployed in practice. To this end, we investigate the ability of an agentic language model feedback framework to generate planning domains from natural language descriptions that have been augmented with a minimal amount of symbolic information. In particular, we evaluate the quality of the generated domains under various forms of symbolic feedback, including landmarks, and output from the VAL plan validator. Using these feedback mechanisms, we experiment using heuristic search over model space to optimize domain quality.
\end{abstract}

\section{Introduction}

AI planning is a sub-area of Artificial Intelligence, where the objective is to synthesize a plan, guaranteed to generate a state containing the desired goals. A known bottleneck of planning is the authoring of accurate and complete planning domain models, formal descriptions of actions, objects, and constraints on which planners operate. The ability for these planning models to be created automatically from natural language descriptions would greatly expand the accessibility of planning techniques. Recently, large language models (LLMs) have shown promise in generating planning domain models from natural language descriptions \cite{DBLP:conf/nips/GuanVSK23, DBLP:conf/nips/MahdaviATC24, DBLP:journals/corr/abs-2409-15915,oswald-et-al-icaps2024}, yet the resulting domains are often syntactically correct but semantically flawed.

A growing body of research augments generation of planning domain models with feedback, iteratively refining the model after an initial domain is generated 
\cite{DBLP:conf/nips/GuanVSK23,gestrin-et-al-icaps2024wshaxp,DBLP:journals/corr/abs-2409-15915}. 
Earlier approaches typically leverage plan validation feedback \cite{DBLP:conf/nips/GuanVSK23}, interact with an environment \cite{DBLP:conf/nips/MahdaviATC24} to provide corrective feedback to the model, or provide an oracle for filtering generated actions \cite{DBLP:journals/corr/abs-2409-15915}.
However, these approaches often rely on a either a single type of feedback or are generating more than just an action model, for example, problems and plans as well, which can lead to compounding errors. Additionally, many of these approaches rely on benchmarks that are limited in scope, focusing on a small number of well-known domains, such as in \citet{DBLP:conf/naacl/ZuoVLLB25}, which may not generalize to more complex or novel domains.

In this work, we investigate the use of symbolic feedback mechanisms to improve the quality of LLM-generated planning domains from natural language description. In particular, we explore two feedback sources: landmarks \cite{hoffmann-et-al-jair2004} and plan validation from VAL \cite{howey-long-icaps2003wscompetition}. Inspired by the model-space search problem  in \citet{caglar-et-al-aaai2024}, we propose a framework that uses heuristic search over the space of possible feedback messages to iteratively refine the generated domains. For an evaluation, we adapt the heuristic domain equivalence (HDE) measure from \citet{oswald-et-al-icaps2024} to automatically assess the quality of the generated domains against ground truth domains and we assess performance on novel domains that are not present in the LLM training data.  Our results show that feedback significantly improve the quality of the generated domains over the baseline. Further, feedback with search method with gpt-5-mini is able to reach correctness at least once for each of the domains. 

\section{Background \& Related Work}

\subsection{Automated Planning}
In this work, we will discuss our method and pipeline in terms of typed lifted STRIPS planning. Our formalization of lifted STRIPS planning problems is a slightly modified version of \citet{Corra2022BestFirstWS} and \citet{DBLP:conf/ijcai/WichlaczH022}. A typed lifted STRIPS planning problem is a 6-tuple $\Pi = \langle \mathcal{T}, \mathcal{F}_\mathcal{X}, \mathcal{C}, \mathcal{A}, s_0, S_{\ast} \rangle$ where:
$\mathcal{T}$ is a set of types equipped with a tree order $\leq$, (a partial order such that for all $t \in T$ the set of all of its predecessors of $t$ is well ordered).
$\mathcal{F}_\mathcal{X}$ is a finite set of lifted predicate symbols with a set of variables $\mathcal{X}$. Each lifted predicate in $f \in \mathcal{F}$ has an associated arity $k$ along with a list of $k$ variables $x_0, \cdots, x_k \in \mathcal{X}$ and types for each variable $t_1,\cdots,t_k \in \mathcal{T}$. 
$\mathcal{C}$ is a set of object symbols representing objects in the world, equipped with a type map $T: \mathcal{C} \to \mathcal{T}$ that maps objects to their type.
We define the set of grounded predicates $\mathcal{F}_\mathcal{C}$ to be the set of predicates $\mathcal{F}$ where each variable $x_i$ with type $t_i$ has been replaced with an object $c_i$ from $\mathcal{C}$, that meets the type constraint: $T(c_i) \leq t_i $. A state $s \subseteq \mathcal{F}_\mathcal{C}$ is a set of grounded predicates representing what is true at a given moment. A state $s \subseteq \mathcal{F}_\mathcal{C}$ is a set of grounded predicates. The set of all possible states $S$ is the power set of $\mathcal{F}_\mathcal{C}$.
$\mathcal{A}$ is a set of \textit{action schema} where each $a \in \mathcal{A}$ is a 4-tuple $\langle \mathcal{X}_a, pre(a), add(a), del(a) \rangle$ where   
$\mathcal{X}_a \subset \mathcal{X}$ is a list of action variables $x_0,\cdots,x_k$, each of which has an associated type $t_1,\cdots,t_k$. 
$pre(a) \subseteq \mathcal{F}_{\mathcal{X}_a}$ is the set of lifted predicates with variables from $\mathcal{X}_a$ that must hold to apply the action, $add(a) \subseteq \mathcal{F}_{\mathcal{X}_a}$ is the set of lifted predicates that become true after the action is applied, and  $del(a) \subseteq \mathcal{F}_{\mathcal{X}_a}$ is the set of lifted predicates that become false after the action is applied. 
Note, for every occurrence of a variable $x_i$ with type $t_i$ in $pre(a)$, $add(a)$, and $del(a)$ it must meet the type constraint for where it appears in the given predicate. 
As with predicates, we ground an action schema by replacing each variable $x_i$ of type $t_i$ with an object $c_i$ that meets the typing requirement $T(c_i)\leq t_i$. We call the set of all grounded actions $\mathcal{A}_g$. Finally, we have the initial state $s_0 \in S$ representing the initial state of the world,
 and $S_* \subseteq S$ the set of goal states. Given a grounded action $a_g \in \mathcal{A_g}$ and a state $s \in S$, we say $a_g$ is {\em applicable} in $s$ iff $pre(a_g) \subseteq s$. Applying an applicable action $a_g$ in the state $s$ results in a new state $s[a_g] := (s / del(a_g)) \cup add(a_g)$. 
A \textit{plan} for a problem $\Pi$ is a sequence of grounded actions $\pi = (a_1, \cdots , a_n)$ which when applied transforms the initial state $s_0$ into a goal state in $S_{\ast}$. The action sequence defines a state sequence $\mathbf{S_\pi} = (s_0, \cdots, s_n)$ such that $s_i = s_{i-1}[a_i]$ for $1\leq i\leq n$ and $s_n\in S_{\ast}$. The set of all plans for $\Pi$ is denoted by $\mathcal{P}_{\Pi}$. A \textit{fact landmark} for a planning problem $\Pi$ is a formula $\phi$ (of conjunctions, disjunctions, negations, etc) over grounded predicates $f_1,\cdots,f_n$ that holds at some point in the state sequences of all plans. We say a formula $\phi$ holds at a state $s \in S$, written $s \vDash \phi$, iff $\phi$ is satisfied after replacing each $f_i$ in $\phi$ with the truth value of $f_i \in s$. Hence, $\phi$ is a state landmark iff $\forall\pi \in \mathcal{P}_{\Pi}, \exists s \in \mathbf{S_\pi}, s \vDash \phi$. A \textit{disjunctive action landmark} $l \subseteq \mathcal{A}_g$ for a planning problem $\Pi$ is a set of grounded actions, at least one of which must hold on all plans, that is $l$ is a disjunctive action landmark iff $\forall \pi \in \mathcal{P}_\Pi, \exists a_g \in l, a_g \in \pi$.     

A \textit{planning domain} for a typed lifted STRIPS planning problem $\Pi$ is the problem's type, predicate, and action schema sets $\mathbf{D} = 
\langle \mathcal{T}, \mathcal{F_\mathcal{X}}, \mathcal{A} \rangle$. We say $\Pi$ is a problem for $\mathbf{D}$ and write $\Pi_\mathbf{D}$ if $\Pi$ uses $\mathbf{D}$ as its underlying domain, regardless of the specific objects, initial state, and goal states ($\mathcal{C}$, $s_0$, $S_{\ast}$) for the problem.

For the representation of planning domains and problems in a human and LLM readable format, this work uses PDDL (Planning Domain Definition Language) \cite{mcdermott-et-al-tr1998}, which structures planning problems in an S-Expression format. Examples of PDDL can be found in the appendix.

\subsection{Domain Model Generation Via Reprompting}

Multiple existing methods examine feedback based approaches to fixing generated PDDL. Originally the approach provided by \citet{DBLP:conf/nips/GuanVSK23} investigated reprompting with feedback from a plan validation tool VAL to fix syntax and semantics errors. We also use VAL. However, their evaluation of generated domain quality was limited by the need for human reviewers. The method presented by \citet{DBLP:conf/nips/MahdaviATC24} generates a PDDL domain and problem from a natural language description of a domain, problem, action interface, and object list. Additionally, their approach makes a strong assumption that for feedback they have access to an oracle that tests the validity
of plans in the ground truth domain. In contrast, we use a weaker assumption and instead provide a set of problems and plans that work the ground truth domain for feedback. \citet{DBLP:journals/corr/abs-2409-15915} presents a method that as part of its pipeline is given a NL description, creates multiple PDDL action candidates for each action, and uses a sentence encoder to filter candidates for valid PDDL domains. Our work differs from these approaches in that we investigate multiple types of symbolic feedback, including landmark feedback, and use heuristic search to guide the feedback selection process. \citet{gestrin-et-al-icaps2024wshaxp} is a full domain, problem, and plan generation pipeline from minimal text descriptions. For their domain construction phase, they use various forms of syntactic feedback from VAL, Loki, and cpddl. In contrast, our work exclusively focuses on generating domains and is interested in investigating the effect of feedback based on problems and plans from the true domain. Similarly to \citet{DBLP:conf/nips/GuanVSK23} and our work, they use VAL for generating feedback on LLM output. 
However, we do not use VAL for syntactic feedback, rather using it for plan feedback. \citet{DBLP:conf/naacl/ZuoVLLB25}, presents a benchmark for text to PDDL generation for problems, and investigates multiple LLM-based approaches to generating PDDL domains from text. Their benchmark, however, only focuses on three small, 
well known and well studied domains, which risks the possibility of models being overfitted to these domains. Our work instead focuses on a wider variety of domains, including some that are novel and have never been seen before by any language models. \citet{casciani2025requirements} presents a method for generating planning domains from natural language descriptions using LLMs, with a focus on requirements engineering and invariant-based refinement. Their approach emphasizes the importance of capturing domain invariants to ensure the generated domains are consistent and reliable. While our work also explores domain generation from natural language, we don't specifically focus on requirements engineering or invariant-based refinement, but rather investigate the use of landmark and plan feedback mechanisms and heuristic search to improve domain quality. \citet{DBLP:conf/acl/HuCZLCL00SL25} presents a benchmark for text to PDDL generation, with a focus on evaluating the performance of LLMs in generating planning domains from natural language descriptions. Their benchmark includes a variety of domains and problem instances. Their feedback mechanism focuses on plan validation and syntactic correctness, similar to some of the feedback mechanisms we explore. However, our work extends this by investigating landmark feedback and heuristic search strategies to guide the feedback process, aiming to improve the overall quality of the generated domains. \citet{DBLP:conf/icra/ZhouSYS024} presents a method for generating PDDL domains, problems, and plans from natural language descriptions using LLMs. Their approach focuses on an iterative refinement process of the plans rather than the domain itself, while our work focuses on their preprocessing stage of domain generation, and investigates multiple feedback mechanisms to improve domain quality. \citet{Yu2025GeneratingSW} looks at a method for PDDL generation from natural language using Chain of thought with LLMs. Their approach focuses on generating PDDL domains and problems from natural language descriptions using a chain-of-thought prompting strategy to guide the LLM in producing more accurate and coherent outputs.

\section{Method}\label{sec:Method}

\begin{figure}
  \centering
  \includegraphics[width=\linewidth]{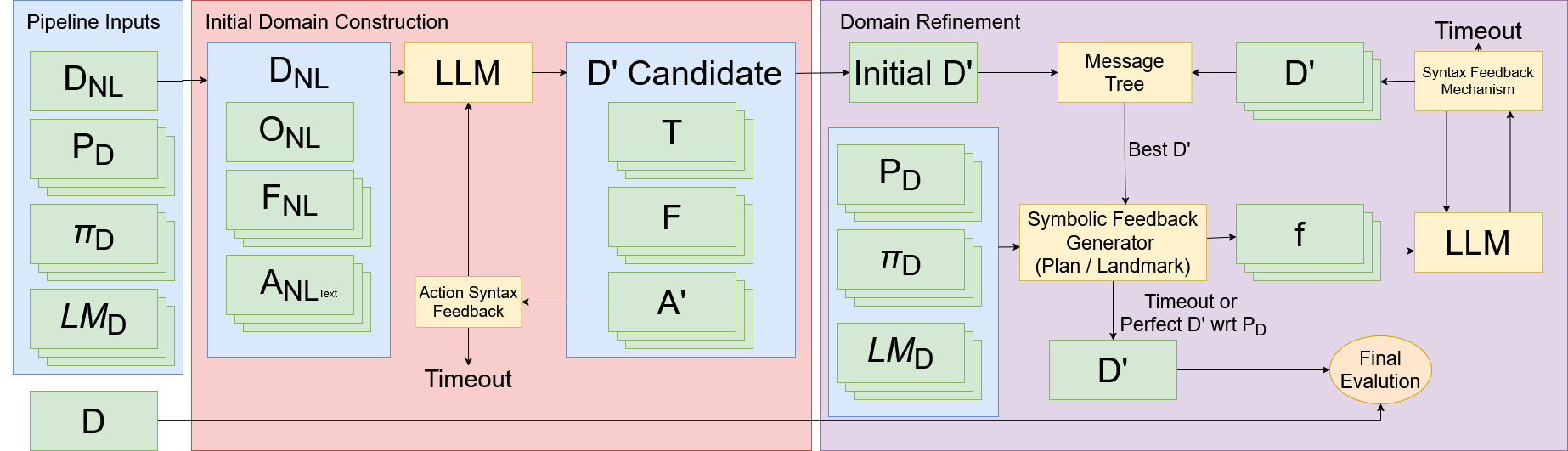}
  \caption{An overview of our model space reasoning as search in feedback space pipeline for planning domain generation from natural language descriptions. 
  }
  \label{fig:pipeline}
\end{figure}

The goal of our method is as follows: Given a natural language description of a planning domain $\mathbf{D}_{NL}$ and a set of auxiliary information for feedback including problems $\mathbf{P}_\mathbf{D}$, plans for those problems $\pi_\mathbf{D}$, and landmarks for these problems $LM_\mathbf{D}$, we wish to use a language model $L$ to generate a high quality PDDL domain $\mathbf{D}'$ that closely models $\mathbf{D}_{NL}$. Additionally, we wish to automatically evaluate the quality of $\mathbf{D}_{NL}$ with respect to $\mathbf{D}'$. To this end, we assumed that $\mathbf{D}_{NL}$ is structured as a triple $(O, \mathcal{F}_{NL}, \mathcal{A}_{NL})$ composed of three parts: (1) $O$, an overall description of the domain, describing the general setting. (2) $\mathcal{F}_{NL}$, a set of natural language descriptions for a subset of predicates in the domain, describing what they mean in the context of the domain. (3) $\mathcal{A}_{NL}$, a set of natural language descriptions for each action in the domain, including its name, and possibly natural language descriptions of its preconditions and effects depending on what type of feedback is being used.
Note that we make no restriction enforcing the description of types, predicates, \footnote{With the exception of the plan feedback pipeline, where the names of initial state predicates and goal predicates present in feedback problems are required due to the need to validate existing feedback plans against the generated domain. We importantly note that due to our results with landmark feedback, which don't require this, we don't particularly need to worry about the assumption being too strong in this case.} or preconditions and effects for actions, but note that describing preconditions and effects arise naturally when describing actions, for examples of these descriptions, see Appendix \ref{appendix:domains}. 

\subsection{Pipeline Overview}
Our generation pipeline is composed of two phases, an initial domain construction phase and a domain refinement phase where the domain is iteratively refined with corrective feedback. We investigate a number of different search-based feedback pipelines, including no feedback (N), landmark feedback with search (LS), plan feedback with search (VS), random single plan feedback, random single landmark feedback (LVR), plan and landmark feedback with search (LVS), and random single plan and landmark feedback. An overview of the pipeline can be seen in Figure \ref{fig:pipeline}.

\subsection{Initial Domain Construction}
Our domain construction phase starts by setting up an initial history $H_0$ composed of an initial system prompt and two context examples converting a natural language description of an action into PDDL. For each action description $a_{NL} \in \mathcal{A}_{NL}$ and querying $L$ to produce a PDDL action $a$, along with a list of predicates $\mathcal{F}$ and types $T$ used in the action, and natural language descriptions of each. The generated $a$ is now checked against a syntactic validators to ensure that the model has produced valid PDDL code for the action. If $a$ is not syntactically valid, an error message about how the PDDL parser failed to parse the action, along with a retry request is appended onto the history and $L$ is called again. This process of syntactic repair for action generation is repeated until we end up with a correct action or we reach a cutoff threshold. After constructing the first action, we save it to a list of completed actions $A$ then move onto constructing the next. For the next action we provide the history $H$ from constructing the previous action and also reiterate the list of types and predicates used so far and tell the model to prefer these and only create new predicates when necessary. Once all $a$ in $A_{NL}$ have been iterated through, we begin a domain construction phase, where the list of types, predicates, and valid actions are put together into an initial PDDL domain $\mathbf{D}'$.  

\subsection{Domain Quality Evaluation}
Once we have a domain $\mathbf{D}'$ we would like to evaluate its quality with respect to $\mathbf{D}_{NL}$. Some previous work such as \citet{DBLP:conf/nips/GuanVSK23,gestrin-et-al-icaps2024wshaxp} use a human evaluation where a person manually checks the quality of $\mathbf{D}'$ against a natural language description or a ground truth domain. In order to remove the need for human evaluators, we use a modified version of the heuristic domain equivalence (HDE) measure from \citet{oswald-et-al-icaps2024}, where generated domains $\mathbf{D}'$ are automatically compared with a ground truth domain $\mathbf{D}$ which has been manually constructed to model the natural language description of the domain $\mathbf{D}_{NL}$. HDE requires not only a ground truth domain $\mathbf{D}$ but also some set of evaluation problems for $\mathbf{D}$, $\mathbf{P}_\mathbf{D}^{eval}$ which are used in HDE computation, providing the ground truth domains and evaluation problems for each of our domains. We note that this ground truth domain and set of associated planning problems for evaluation is never given to the model, and exclusively used for evaluation. We further note that $\mathbf{P}_\mathbf{D}^{eval}$, the set of evaluation problems, is distinct from the set of feedback problems seen in Figure \ref{fig:pipeline}, $\mathbf{P}_\mathbf{D}$.
Heuristic domain equivalence is computed as follows: given a ground truth planning domain $\mathbf{D}$, a generated planning domain $\mathbf{D}'$, and a set of solvable planning problems for $\mathbf{D}$, $\mathbf{P}_\mathbf{D}^{eval}$, each problem $\Pi\in \mathbf{P}_\mathbf{D}^{eval}$ can be transformed into a problem $\Pi'\in \mathbf{P}_{\mathbf{D}'}$ that uses $\mathbf{D}'$ as its underlying domain by replacing the domain of the the problems. This works since our construction of $\mathbf{D}'$ guarantees that action names and the subset of predicate names in the problems will match.
For each such pair of problems $\Pi$ and $\Pi'$, we take corresponding subsets of their plans $P\subseteq \mathcal{P}_{\Pi}$ and $P'\subseteq \mathcal{P}_{\Pi'}$ and we can cross check whether $P\subseteq\mathcal{P}_{\Pi'}$ and cross check their validity. We define and use a normalized version of the HDE measure formally as follows, where the second term is taken to be $0$ if $|P'|=0$:

\begin{equation*}
  \text{HDE}(\Pi,\Pi') = \frac{1}{2}\left(\frac{|P \cap \mathcal{P}_{\Pi'}|}{|P|} + \frac{|P' \cap \mathcal{P}_{\Pi}|}{|P'|} \right)
\end{equation*}

The first term in the HDE measures the number of plans that pass in the forward direction, i.e., how many of the ground truth plans are valid in the generated domain, while the second term measures the number of plans that pass in the backward direction, representing how many of the generated plans that work in the new domain are valid in the ground truth domain. While at first glance these may seem to measure the same thing, the directionality captures an important notion about the failure mode. Note that $|P| \neq |P'|$ in general, in situations where $|P| << |P'|$ it is possible to have a high forward HDE score but a low backward HDE score, indicating that the generated domain may be overgeneralized with respect to the ground truth domain, where as the opposite situation could be an indicator that the generated domain is too restrictive. For a final HDE score between domains $\mathbf{D}$ and $\mathbf{D}'$ we take the average HDE score across all problem pairs $\Pi \in \mathbf{P}_\mathbf{D}$ and their corresponding $\Pi' \in \mathbf{P}_{\mathbf{D}'}$.
In practice to generate subsets of plans $P$ and $P'$, we use a top-k planner, K$^*$, \cite{lee-et-al-socs2023}. To compute $P \cap \mathcal{P}_{\Pi'}$ and $P' \cap \mathcal{P}_{\Pi}$ for each individual plan, the test can be efficiently performed using a plan validator. For this work we use VAL.\footnote{\url{https://github.com/KCL-Planning/VAL}}

\subsection{No Feedback Pipeline}
To experiment with various types of feedback, we introduce the ``No Feedback'' pipeline as a baseline where no feedback is given. This pipeline takes the domain $\mathbf{D}'$ that comes out of the domain construction phase and immediately evaluates it. Note that $\mathbf{D}'$ is guaranteed to be syntactically and semantically correct by construction when it exits the domain construction phase.

\subsection{Syntax and Semantics Feedback Mechanism}
Across all of the following pipelines, we use feedback mechanisms on the output of any domain $\mathbf{D}'$ after it exits the language model to generate feedback on the syntactic and semantic correctness of $\mathbf{D}'$. In particular, we use a PDDL parser\footnote{\url{https://github.com/AI-Planning/pddl}} to ensure that $\mathbf{D}'$ is syntactically valid PDDL. If it is not, a feedback error message is produced, and the model is re-prompted to fix the syntax errors, until none remain or the pipeline times out. This is necessary as our other forms of feedback assume that $\mathbf{D}'$ is a syntactically valid PDDL domain. It should be noted that this isn't necessary for the No Feedback Pipeline, as $\mathbf{D}'$ is guaranteed to be syntactically valid when exiting the model. It should also be noted that the syntactic feedback mechanism does not provide any feedback to the model about the quality of $\mathbf{D}'$ with respect to $\mathbf{D}_{NL}$.

\subsection{Plan Feedback}

For random-single plan feedback, the model is provided with a list of feedback plans $\mathbf{P}_\mathbf{D}$ which are valid plans with respect to the ground truth domain $\mathbf{D}$ that $\mathbf{D}_{NL}$ models. For feedback each $\pi \in \mathbf{P}_\mathbf{D}$ is evaluated for validity against the ground truth problems $\Pi_\mathbf{D}$ mapped to use the generated domain $\mathbf{D}'$ instead of the ground truth domain $\mathbf{D}$, i.e., each plan $\pi \in \mathbf{P}_D$ is checked with respect to $\Pi_\mathbf{D}'$ (using VAL). Each invalid feedback plan produces a single feedback message, which may include two types of errors (for an example of a plan feedback message see Appendix \ref{apx:plan_feedback_prompt}). The first type of error is when an action $a' \in \pi$ is not applicable to the current state during validation, indicating that the preconditions of $a'$ are not being met, which could be due to an incorrect precondition on $a'$ or missing effects on the previous actions, etc. The second type of error is when the plan ends in a state that does not satisfy the goal conditions, indicating that some action effects may be incorrect or missing. Once this feedback is generated, the random single pipeline will select a single feedback message at random and append it to the history, then re-query the model to produce a new domain $\mathbf{D}'$. This process is repeated until no invalid plans remain in $\mathcal{P}_\mathbf{D}$, in which case we proceed to a final evaluation or the timeout threshold is reached.

\subsection{Landmark Feedback}

For landmark feedback on $\mathbf{D}'$ we use the set of disjunctive action landmarks $LM_\mathbf{D} = \{a_11 \lor a_12\cdots,a_21 \lor a_22\cdots, \cdots\}$ generated by \texttt{forbiditerative} from \citet{katz-sohrabi-aaai2020} as the main way we provide feedback to the model. These landmarks are generated from the ground truth domain $\mathbf{D}$ and a set of problems $\Pi_D$. 
To actually generate feedback, we begin by using the generated domain $\mathbf{D}'$ and our list of problems $\Pi_\mathbf{D}$ to create a set of problems $\Pi_\mathbf{D}'$ that use $\mathbf{D}'$ as their underlying domain. We then use a top-k planner, K$^*$ \cite{lee-et-al-socs2023} to generate a set of plans $P'$ for each problem in $\Pi_\mathbf{D}'$. For an example landmark prompt see Appendix \ref{apx:landmark_feedback_prompt}.
For each action landmark $a_1\lor\cdots\lor a_n \in LM_\mathbf{D}$ and each plan $\pi \in P'$  we check if at least one action from $a$ appears in any of the plans generated for the problems $\Pi_\mathbf{D}'$ using the generated domain $\mathbf{D}'$. If no action from $l$ appears in any of the plans, a feedback message is generated indicating that at least one action from $l$ must appear the given plan. Similar to the random-single plan feedback pipeline, a single feedback message is selected at random and appended to the history, and the model is re-queried to produce a new domain $\mathbf{D}'$. This process is repeated until no invalid landmarks remain or the timeout threshold is reached.

\subsection{Random-Single Plan and Landmark Feedback}

This pipeline combines the random-single plan feedback and random-single landmark feedback pipelines. In particular, both forms of feedback are generated for the current domain $\mathbf{D}'$, and a single feedback message is selected at random from the combined set of feedback messages and appended to the history. The model is then re-queried to produce a new domain $\mathbf{D}'$. This process is repeated until no invalid plans or landmarks remain or the timeout threshold is reached.

\subsection{Heuristic Search Pipelines}

For each of the feedback mechanisms presented in the random single pipelines (LR, VR, LVR), we also investigate the use of a heuristic search over the space of possible feedback messages to help select which feedback to provide to the model for improving domain quality. Specifically, we use a best-first search heuristic over evaluation on the feedback problems. We implement this approach using a tree structure to keep track of the message history to the model and the various domains $\mathbf{D}'$ that have been generated so far, where each node in the tree represents a domain $\mathbf{D}'$, and its children represent the domains generated by providing different feedback messages to the model. The root node of the tree is the initial domain $\mathbf{D}'$ that comes out of the domain construction phase. The leaves of the tree are the open list of domains that have not yet been expanded with feedback. Once a domain $\mathbf{D}'$ is selected from the open list, we select $n$ possible feedback messages ($f$ in Figure \ref{fig:pipeline}) for it (where $n$ is a hyperparameter, if $n=1$, search functions identically to random-single, if $n=0$ search functions identically to to the no feedback pipeline), and each feedback message in $f$ is used to re-query the model to produce a set of new domains $\mathbf{D}''_1, \mathbf{D}''_2, \ldots, \mathbf{D}''_n$, which become child nodes of $\mathbf{D}'$, thus being added to the open list. Once nodes are generated, they are scored using a heuristic function based on $G$, the depth of the node in the tree, and $H$, the number of invalid plans in the domain $\mathbf{D}'$, i.e., $H = |P| - |P \cap \mathcal{P}_{\Pi'}|$, where $P$ is the set of feedback plans. Some hyperparameters are used to control the behavior of the search, including a weight for $H$, as well as a positive integer that caps the number of possible feedback messages that can be generated for a given domain $\mathbf{D}'$. The search continues until a domain is found with no invalid plans ($H = 0$), or the timeout threshold is reached.

\section{Dataset}
For our dataset we evaluate the PDDL domain generation ability of a wide range of typed STRIPS PDDL domains. An overview can be seen in Appendix \ref{appendix:dataset} Table \ref{tab:domains}. Some of our domains, such as {\tt blocks} and {\tt miconic} are classic planning domains found in the international planning competition \cite{mcdermott-et-al-tr1998}, and have large amounts of papers and popular resources written about them available on the web such as \cite{slaney-thiebaux-aij2001,russell-norvig-2003}. Others, such as {\tt hiking}, {\tt pacman-63}, and {\tt pacman-72} are obscure domains that appear in just a few papers or public repositories on the web \cite{DBLP:conf/acl/HuCZLCL00SL25}. The rest are novel domains that have been designed specifically for this work, or were designed by us before model training data cutoff dates. Each domain in our dataset is equipped with a problem generator for generating a set of solvable problems for the domain. 

\begin{figure*}
  \centering
  \includegraphics[width=0.99\linewidth]{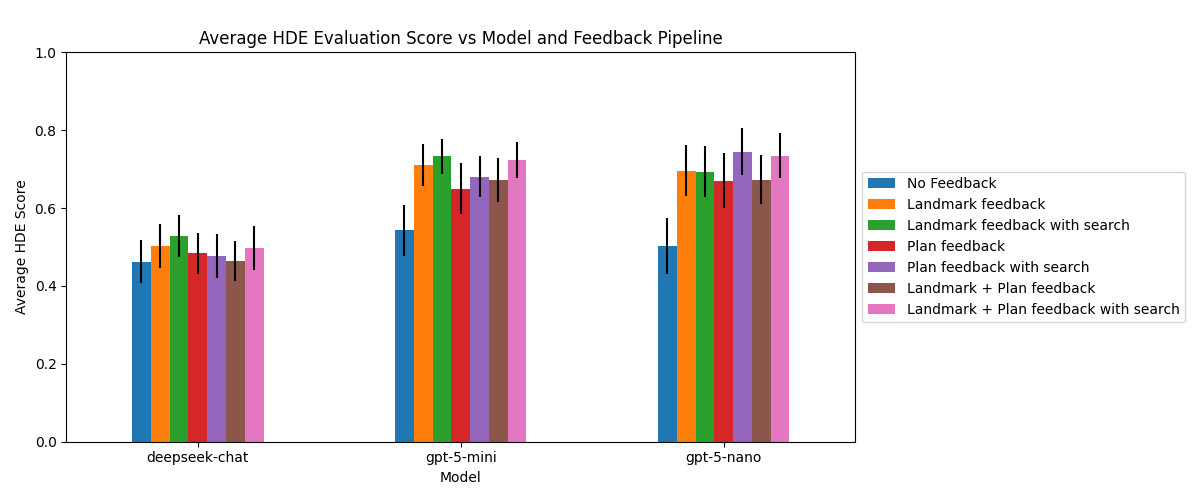}
  \caption{HDE score averaged across all domains for each model and feedback pipeline. Error bars indicade 95\% confidence intervals over trials, i.e., 95\% of trials scored within the error bars.}
  \label{fig:results}
\end{figure*}

\section{Experiments}

For our experimental evaluation, we are trying to answer the following research questions about our model space reasoning as search in feedback space pipeline for planning domain generation from natural language descriptions: 
%
\begin{enumerate}[start=1,label={ \bfseries R\arabic*.},ref={R\arabic*},leftmargin=*, noitemsep, topsep=0pt]
\item Does feedback meaningfully improve domain quality over the no feedback baseline? \label{R:nofeedback}
\item How is the quality of generated PDDL domains from natural language descriptions affected by various forms of feedback?  \label{R:qualityfeedback}
\item Does combining multiple forms of feedback lead to further improvements? \label{R:multiplefeedback}
\item Does a systematic search in feedback space consistently lead to better outcomes compared to a random walk? \label{R:feedbacksearch}
\end{enumerate}
To this end we set up a large grid of experiments varying the domain $\mathbf{D}_{NL}$, language model $L$, feedback pipeline, and description classes (variations of $\mathcal{F}_{NL}$ and $\mathcal{A}_{NL}$). Each experiment is repeated across multiple trials n=20 with different random seeds and descriptions to account for variability in model outputs.

\subsection{Experiment Setup} 
All of our experimental evaluations 
were done on a system with an AMD Ryzen 9 7950X 16-Core Processor and 64GB of memory running Ubuntu 22.04 LTS. 
The core experimental parameters we vary over are the choice of language model and feedback pipeline. For the language model we evaluate over \texttt{gpt-5-nano}, \texttt{gpt-5-mini}, and \texttt{deepseek-chat}. For our experiment pipelines we evaluate over seven different methods. (1) the baseline, \textbf{No feedback} (N) \cite{oswald-et-al-icaps2024}, where we use the domain $\mathbf{D}'$ directly from initial domain construction phase without any refinement. (2) \textbf{Landmark Feedback} (LR) where a single random landmark feedback message is given to the model at each step to generate a series of successive $\mathbf{D}'$s until there is no feedback left to give, or we reach a timeout threshold. (3) \textbf{Landmark Feedback with Search} (LS), where we select the best successive $\mathbf{D}'$ according to heuristic search on domains resulting from multiple pieces of landmark feedback. (4) \textbf{Plan Feedback} (VR), where a single random VAL (plan) feedback message is given to the model at each step to generate a series of successive $\mathbf{D}'$s until there is no feedback left to give, or we reach a timeout threshold. (5) \textbf{Plan Feedback with Search} (VS), where we select the best successive $\mathbf{D}'$ according to heuristic search on domains resulting from multiple pieces of plan feedback. (6) Landmark + Plan Feedback (LVR), where a single random landmark or plan feedback message is given to the model at each step to generate a series of successive $\mathbf{D}'$s until there is no feedback left to give, or we reach a timeout threshold. (7) \textbf{Landmark + Plan Feedback with Search} (LVS), where we select the best successive $\mathbf{D}'$ according to heuristic search on domains resulting from multiple pieces of plan feedback and landmark feedback.
A number of experimental hyperparameters were fixed for our experiment. We start with $\mathbf{D}$, a ground truth domain and handcrafted $\mathbf{D}_{NL}$. We provide two different descriptions for each for each $\mathbf{D}_{NL}$ which can be seen in the description in Appendix \ref{appendix:domains}. Using $D$ and a problem generator for $\mathbf{D}$ we generate a set of solvable feedback problems $\mathbf{P}_\mathbf{D}$ and a set of solvable evaluation problems $\mathbf{P}_\mathbf{D}^{eval}$. For our experiments, we fix the size of the feedback problem set to 5 and the size of the evaluation problem set to 5. For each problem in the feedback problem set $\Pi \in \mathbf{P}_\mathbf{D}$ we generate at most two feedback plans, $\pi_\mathbf{D}$. For each problem in the evaluation problem set $\Pi \in \mathbf{P}_\mathbf{D}^{eval}$ at most one hundred evaluation plans, $\pi_\mathbf{D}^{eval}$. For our threshold in generating actions, we give the model five chances to generate a syntactically valid action before aborting. For all pipelines that use refinement, we set the timeout threshold to 15, which includes both syntax feedback and symbolic feedback steps. i.e., LR, VR, LVR are bounded to a message maximum depth of 15, while LS, VS, LVS can expand up to 15 nodes in the open list. In the LS, VS, LVS pipelines, when expanding a node on the open list we limit our expansions to 10 nodes, IE, we cap the number of feedback messages passed to the LLM at 10. In practice, since there are only two feedback plans  this never matters for VR or VS. For landmark feedback we observe we rarely pass this threshold, occurring in less than 10\% of trials. 
We note that for purposes of analyzing the Table \ref{tab:full_results}, there are 20 trials in total, we run 10 runs over randomized LLM seeds for both of our description classes.

\vspace{-0.3cm}

\subsection{Results}

\begin{table*}[t] 
  \centering
\setlength\tabcolsep{1.2pt} 
\def\arraystretch{1.02}
  \begin{tabular}{l|r|rrrrrrr|r||rrrrrrr}
& & \multicolumn{8}{|c||}{Average HDE \%} & \multicolumn{7}{c}{\# Perfect @ 20} \\
\hline
 & Domain & N & LR & LS & VR & VS & LVR & LVS & Avg & N & LR & LS & VR & VS & LVR & LVS \\ 
\hline
\multirow{9}{*}{\rotatebox{90}{deepseek-chat}}& blocks & 59±5 & 54±5 & \bfseries 66±5 & 52±4 & 51±5 & 59±5 & 59±4 & 57±5  & 1 & 1 & \bfseries 2 & 0 & 1 & 0 & 0\\
& bloxorz & 31±7 & \bfseries 42±9 & 41±9 & 28±7 & 39±8 & 33±7 & 36±8 & 36±8  & 1 & \bfseries 3 & \bfseries 3 & 0 & 2 & 1 & 1\\
& checkers & 1±1 & 14±5 & 8±4 & 15±5 & 5±3 & 15±5 & 3±3 & \bfseries 9±4  & \bfseries 0 & \bfseries 0 & \bfseries 0 & \bfseries 0 & \bfseries 0 & \bfseries 0 & \bfseries 0\\
& flow & 33±8 & 40±10 & \bfseries 55±9 & 43±7 & 42±8 & 29±8 & 37±9 & 40±8  & 1 & \bfseries 4 & \bfseries 4 & 1 & 1 & 1 & 3\\
& hiking & 68±5 & 71±5 & 73±5 & 79±6 & \bfseries 85±4 & 70±5 & 76±5 & 75±5  & 4 & 4 & 7 & \bfseries 9 & 8 & 5 & 6\\
& miconic & 65±4 & 66±5 & 67±4 & \bfseries 69±4 & 64±3 & \bfseries 69±4 & 68±4 & 67±4  & 3 & \bfseries 4 & 3 & 2 & 1 & 3 & 3\\
& pacman-63 & 24±5 & 19±4 & \bfseries 31±4 & 20±4 & 14±4 & 23±5 & 27±5 & 23±4  & \bfseries 0 & \bfseries 0 & \bfseries 0 & \bfseries 0 & \bfseries 0 & \bfseries 0 & \bfseries 0\\
& pacman-72 & 72±4 & \bfseries 82±4 & 62±3 & 69±4 & 67±4 & 60±3 & 75±5 & 70±4  & 3 & \bfseries 6 & 1 & 3 & 4 & 1 & 5\\
\hline
& Aggregate & 44±5 & 48±6 & \bfseries 51±6 & 47±5 & 46±5 & 45±5 & 48±5 & 47±5  & 13 & \bfseries 22 & 20 & 15 & 17 & 11 & 18\\
\hline
\multirow{9}{*}{\rotatebox{90}{gpt-5-nano}}& blocks & 46±10 & 47±11 & 56±10 & 52±11 & \bfseries 67±10 & 43±10 & 65±9 & 54±10  & 3 & 4 & 5 & 5 & \bfseries 6 & 3 & 5\\
& bloxorz & 59±8 & 81±5 & 88±5 & 71±8 & 88±4 & 83±4 & \bfseries 91±4 & 80±6  & 6 & 7 & \bfseries 11 & 6 & 10 & 6 & 10\\
& checkers & 4±3 & 46±10 & 38±6 & 22±10 & 45±6 & 32±5 & \bfseries 50±0 & 34±6  & 0 & 0 & 0 & \bfseries 2 & 0 & 0 & 0\\
& flow & 0±0 & 35±8 & 26±8 & 31±8 & 31±7 & \bfseries 48±7 & 38±7 & 30±6  & 0 & \bfseries 1 & 0 & \bfseries 1 & 0 & \bfseries 1 & \bfseries 1\\
& hiking & 65±7 & 79±8 & 88±6 & 87±4 & \bfseries 95±3 & 91±3 & 78±8 & 83±6  & 4 & 9 & 10 & 10 & \bfseries 13 & 11 & 10\\
& miconic & 75±6 & 73±6 & 71±6 & 75±5 & 69±7 & 65±7 & \bfseries 79±6 & 73±6  & \bfseries 8 & \bfseries 8 & 6 & 7 & 6 & 5 & 7\\
& pacman-63 & 37±5 & 36±4 & 37±4 & 36±5 & 41±4 & 32±5 & \bfseries 46±2 & 38±4  & \bfseries 0 & \bfseries 0 & \bfseries 0 & \bfseries 0 & \bfseries 0 & \bfseries 0 & \bfseries 0\\
& pacman-72 & 81±7 & 92±4 & 87±4 & \bfseries 93±4 & 87±6 & 91±3 & 81±7 & 88±5  & 9 & \bfseries 13 & 10 & \bfseries 13 & 11 & 11 & 10\\
\hline
& Aggregate & 46±6 & 61±7 & 61±6 & 58±7 & 65±6 & 61±6 & \bfseries 66±5 & 60±6  & 30 & 42 & 42 & 44 & \bfseries 46 & 37 & 43\\
\hline
\multirow{9}{*}{\rotatebox{90}{gpt-5-mini}}& blocks & 80±4 & 71±5 & 77±3 & 76±5 & \bfseries 80±5 & 74±5 & 72±6 & 76±5  & 4 & 3 & 2 & 4 & \bfseries 5 & 4 & \bfseries 5\\
& bloxorz & 40±7 & 73±8 & 77±7 & 69±9 & 80±7 & 72±7 & \bfseries 82±7 & 70±7  & 0 & 6 & 4 & 6 & \bfseries 7 & 5 & \bfseries 7\\
& checkers & 42±4 & 46±6 & 55±6 & 43±7 & 50±5 & 54±6 & \bfseries 68±6 & 51±6  & 0 & 1 & 2 & 1 & 1 & 2 & \bfseries 4\\
& flow & 42±9 & \bfseries 87±3 & 65±4 & 45±9 & 72±5 & 62±8 & 74±5 & 64±6  & 2 & \bfseries 6 & 1 & 2 & 4 & 4 & 5\\
& hiking & 75±8 & 78±5 & 82±4 & \bfseries 90±4 & 67±5 & 77±5 & 78±5 & 78±5  & 7 & 6 & 4 & \bfseries 8 & 3 & 5 & 6\\
& miconic & 72±4 & 82±5 & 92±4 & 76±5 & 77±5 & \bfseries 93±4 & 70±5 & 80±4  & 2 & 6 & 9 & 5 & 4 & \bfseries 10 & 3\\
& pacman-63 & 19±5 & 48±2 & 48±1 & 33±4 & 41±4 & 40±4 & \bfseries 55±3 & 40±3  & 0 & 0 & 0 & 0 & 0 & 0 & \bfseries 1\\
& pacman-72 & 65±4 & \bfseries 83±5 & 81±4 & 83±4 & 74±5 & 65±4 & 73±4 & 75±4  & 2 & \bfseries 7 & 5 & 5 & 4 & 2 & 3\\
\hline
& Aggregate & 54±6 & 71±5 & \bfseries 72±4 & 64±6 & 68±5 & 67±5 & 71±5 & 67±5  & 17 & \bfseries 35 & 27 & 31 & 28 & 32 & 34\\
\hline
  \end{tabular}
  \vspace*{0.3cm}
  \caption{HDE results in percentages across all domains and models, averaged over twenty trials varying model seed and description class, reported with standard error on the mean. \# perfect \@ 20 is the number of perfect HDE scores of our 20 trials, i.e the count of HDE scores = 100\%. The ``Aggregate'' row is an average for Average HDE columns and Sum for \# perfect \@ 20 Columns. Columns from left to right: Model name; Domain name; N: No feedback; LR: Landmark feedback with random single feedback; LS: Landmark feedback with search; VR: Plan validation feedback with random single feedback; VS: Plan validation feedback with search; LVR: Landmark and plan validation feedback with random single feedback; LVS: Landmark and plan validation feedback with search; Average: HDE averages across all feedback types and the total number of perfects for that domain across all feedback types. Bold values indicate the best average performing feedback type for each domain and model combination.}\label{tab:full_results}
\end{table*}

Figure \ref{fig:results} presents the aggregated results for our evaluated methods, comparing various feedback forms to the baseline of no feedback \cite{oswald-et-al-icaps2024}, across 3 language models. 
Our first conclusion positively answers question ~\ref{R:nofeedback}, indeed feedback shows a meaningful improvement over the baseline, across feedback types and models. 
To answer the other questions, we take a deeper look at the per-domain results depicted in Table \ref{tab:full_results}. We view the results in the table as aggregation over 20 runs. The left part of the table reports mean and standard error of the HDE score across these runs. The right part counts the number of runs that reached 100\% HDE. 
Our first observation is that the method that integrates both feedback types with search (LVS) with gpt-5-mini successfully obtains a PDDL domain model that scores 100\% HDE at least once for each of the tested domains. This is an important result, as the HDE score is a strong indicator of the generated PDDL domain correctness.
To answer~\ref{R:qualityfeedback}, we observe that there is no single feedback type that dominates the other ones. They seem to have complementary strengths, each excelling on at least one domain. Combining V and L typically, but not always improves over the single feedback type. Interestingly, the decrease in performance when combining feedback types can be quite large, as in, e.g., {\em hiking} domain. This gives us an answer to~\ref{R:multiplefeedback}. Finally, for~\ref{R:feedbacksearch}, systematic search typically works better than random walks. However, there are quite a few exceptions. Looking at the results for gpt-5-mini, the average HDE score is significantly reduced on, e.g., {\em flow} domain (22\% from LR to LS), {\em hiking} (23\% from VR to VS), {\em miconic} (23\% from LVR to LVS). The number of times a perfect HDE score is obtained in these cases is also in favor of the random walk.
In conclusion, the proposed methods in this work seem to be complementary in their performance. They allow us to generate PDDL domains of significantly higher quality than prior work, as measured by HDE score. 

\section{Conclusion and Future Work}
In this work we presented a novel framework for generating PDDL planning domains from natural language descriptions using large language models and a feedback-driven refinement process. We investigated a number of different feedback mechanisms, including landmark feedback and plan validation feedback, as well as different strategies for prioritizing feedback using heuristic search. Our experimental results demonstrated that feedback mechanisms significantly improve the quality of generated PDDL domains compared to a no feedback baseline, and that different models respond differently to various types of feedback. In our experiments, we were able to generate at least one PDDL domain with 100\% HDE score for each tested domain.
We also found that simple landmark feedback can be as effective as more detailed plan validation feedback, which has important implications for making PDDL generation more accessible to non-experts.

In future work, we plan to explore additional forms of feedback, such as invariants \cite{alcazar-torralba-icaps2015} based feedback.
Some feedback types may induce search spaces with large branching factors, which may require using different search strategies for prioritizing feedback. We also plan to conduct user studies to evaluate the usability of our framework for non-expert users, and to explore applications of our approach in real-world planning scenarios.

\bibliographystyle{iclr2026_conference}
\bibliography{citations,abbrv,literatur,crossref}

\newpage
\appendix
\section{Sample PDDL Domain and Description}\label{appendix:domains}
Here we provide our version of the classic blocksworld domain, referred to in Tables \ref{tab:domains} and \ref{tab:full_results} as ``blocks''.

\begin{verbatim}
(define (domain blocks)
    (:requirements :strips :typing)
    (:types block)
    (:predicates 
        (on ?x - block ?y - block)
        (ontable ?x - block)
        (clear ?x - block)
        (handempty)
        (holding ?x - block)
    )
    (:action pick-up
        :parameters (?x - block)
        :precondition (and (clear ?x) (ontable ?x) (handempty))
        :effect (and 
            (not (ontable ?x)) (not (clear ?x)) (not (handempty)) 
            (holding ?x)
        )
    )
    (:action put-down
        :parameters (?x - block)
        :precondition (holding ?x)
        :effect (and 
            (not (holding ?x)) 
            (clear ?x) (handempty) (ontable ?x)
        )
    )
    (:action stack
        :parameters (?x - block ?y - block)
        :precondition (and (holding ?x) (clear ?y))
        :effect (and 
            (not (holding ?x)) (not (clear ?y)) 
            (clear ?x) (handempty) (on ?x ?y)
        )
    )
    (:action unstack
        :parameters (?x - block ?y - block)
        :precondition (and (on ?x ?y) (clear ?x) (handempty))
        :effect (and 
            (not (clear ?x)) (not (handempty)) (not (on ?x ?y))
            (holding ?x) (clear ?y)
        )
    )
)
\end{verbatim}

A sample natural language description of the above blocksworld domain, $\mathbf{D}_{NL}$, given in json would be something like the following. Note as we mention in Section \ref{sec:Method}, $\mathbf{D}_{NL}$ is structured as a triple $(O, \mathcal{F}_{NL}, \mathcal{A}_{NL})$, which can explicitly be seen here, startling with the overall description, followed by predicate descriptions and action descriptions. 

\begin{lstlisting}[numbers=none]
{
    "overall": {
        "simple": "The blocks domain involves stacking and unstacking blocks on a table. The goal is to arrange the blocks in certain configurations by picking up blocks, putting them down on the table, or stacking them on other blocks.",
        "detailed": "This domain models a robot arm manipulating blocks on a table. Each block can be either on the table, on top of another block, or held by the robot's arm. The robot can only hold one block at a time, and can only pick up a block if it has nothing on top of it. The robot needs to carefully plan a sequence of pick-up, put-down, and stack operations to achieve desired block configurations. This planning challenge emphasizes spatial reasoning and requires understanding constraints like a block cannot be moved if another block is on top of it."
    },
    "predicates": {
        "on": {
            "simple": "?x is on top of another block ?y.",
            "detailed": "This predicate indicates that block ?x is physically placed directly on top of another block ?y, creating a stack. The block on top ?x is supported by the block ?y beneath it."
        },
        "ontable": {
            "simple": "The block ?x is on the table.",
            "detailed": "This predicate indicates that a block ?x is placed directly on the table surface rather than on top of another block. The table serves as the bottom support for all block configurations."
        },
        "clear": {
            "simple": "The block ?X has nothing on top of it.",
            "detailed": "This predicate indicates that a block ?x has no other blocks stacked on top of it, making it accessible to be picked up by the robot arm. Only blocks with this property can be moved."
        },
        "handempty": {
            "simple": "The robot's hand is empty.",
            "detailed": "This predicate indicates that the robot's hand is not currently holding any block, making it available to pick up a new block. The robot can only hold one block at a time."
        },
        "holding": {
            "simple": "The robot is holding the block ?x.",
            "detailed": "This predicate indicates that the robot has picked up a specific block and is currently holding it in its hand. While holding a block, the robot cannot pick up any other blocks."
        }
    },
    "actions": {
        "pickup": {
            "simple": "The robot picks up a block ?x from the table.",
            "detailed": "This action allows the robot to lift a block from the table and hold it in its hand. The action can only be performed if the block is on the table, nothing is stacked on top of it, and the robot's hand is currently empty."
        },
        "putdown": {
            "simple": "The robot puts down the block ?x it is holding onto the table.",
            "detailed": "This action allows the robot to place a block ?x it is currently holding onto the table surface. After performing this action, the block will be on the table and the robot's hand will be empty again."
        },
        "stack": {
            "simple": "The robot stacks a block it is holding ?x onto another block ?y.",
            "detailed": "This action allows the robot to place a block it is currently holding ?x on top of another block ?y. The target block ?y must be clear (have nothing on top of it). After performing this action, the robot's hand will be empty."
        },
        "unstack": {
            "simple": "The robot picks up a block ?x from on top of another block ?y.",
            "detailed": "This action allows the robot to lift a block ?x that is currently stacked on top of another block ?y. The block being picked up must be clear (have nothing on top of it), and the robot's hand must be empty before performing this action."
        }
    }
}
\end{lstlisting}

For the rest of our domains, see the dataset in the repository at: \textbf{Repository link redacted for anonymity. }

\section{Prompts}

For system prompts and context examples, the repository at: \textbf{Repository link redacted for anonymity.}

\subsection{Landmark Feedback Prompt}\label{apx:landmark_feedback_prompt}
Landmark feedback prototypically takes the following form, and is inserted into the chat history directly following the output of some domain. \texttt{\{problem\}} is a problem $p \in \mathbf{P}_\mathbf{D'}$ and \texttt{\{plan\}} is a plan $\pi$ that K$^*$ generated for this problem. \texttt{\{landmark\}} is a comma separated string representation of the disjunctive action landmark $l$ that did not have one of its actions execute on $\pi$.
\begin{lstlisting}[numbers=none]
Given the above generated domain, attempting to use it with the following problem:
```
{problem}
```
We expected that the one of the following actions: 
```
{landmark} 
```
would be executed on the following plan, since these actions are a disjunctive action landmark for the problem:
```
{plan}
```
Please revise the previous domain to fix the issue. You may create new predicates and types if needed, but make sure to update the predicate and type lists accordingly. You may not add new requirements to the domain, your output should exclusively be a typed STRIPS domain.
\end{lstlisting}

\subsection{Plan Feedback Prompt}\label{apx:plan_feedback_prompt}
Plan feedback prototypically takes the following form, where \texttt{\{problem\}} is a problem $p \in \mathbf{P}_\mathbf{D'}$ and \texttt{\{plan\}} is a plan $\pi$ in $\pi_D$ that was given for this problem, unlike the plan for landmark feedback which is generated for the problem, here we are checking if a know correct plan works for the problem on the new domain.
\begin{lstlisting}[numbers=none]
Given the above generated domain, attempting to use it with the following problem:
```
{problem}
```
An issue was encountered with the following plan:
```
{plan}
```
The output of the plan validator VAL is:
```
{val_output}
```
Please revise the previous domain to fix the issue. You may create new predicates and types if needed, but make sure to update the predicate and type lists accordingly. You may not add new requirements to the domain, your output should exclusively be a typed STRIPS domain.
\end{lstlisting}

\section{Dataset Domain Composition}\label{appendix:dataset}
\begin{table}[ht!]
\centering
\small
\begin{tabular}{|l|c|c|c|c|c|c|c|c|}
\hline
\textbf{Domain} & 
\textbf{$|\mathcal{A}|$}& 
\textbf{$|\mathcal{F}|$}& 
\textbf{$|\mathcal{T}|$} & 
\multicolumn{2}{|c|}{\textbf{$\mathcal{A}$ arity}} & 
\multicolumn{2}{|c|}{\textbf{$\mathcal{F}$ arity}} & 
\textbf{Del.} \\
\cline{5-8}
 &  &  &  & $<$ & $>$ & $<$ & $>$ &  \\
\hline
blocks            & 4  & 5  & 1 & 1 & 2 & 0 & 2 & yes \\
bloxorz$^*$       & 3  & 4  & 3 & 5 & 7 & 2 & 3 & yes \\
checkers$^*$ & 4 & 6 & 2 & 3 & 5 & 1 & 2 & yes \\
flow              & 3  & 8  & 2 & 2 & 3 & 0 & 2 & yes \\
hiking            & 2  & 6  & 1 & 2 & 2 & 1 & 2 & yes \\
miconic           & 4  & 8  & 2 & 2 & 2 & 1 & 2 & yes \\
pacman-63         & 2  & 7  & 1 & 2 & 2 & 1 & 2 & no  \\
pacman-72         & 2  & 6  & 1 & 1 & 2 & 1 & 2 & yes \\
\hline
\end{tabular}
\vspace*{0.3cm}
\caption{List of domains with some basic properties. Columns from left to right: domain name; number of lifted actions and lifted predicates; number of types; the min. and max. arity of the actions; the min. and max. arity of the predicates; and whether a literal can be a delete effect. A $\cdot^*$ in the name indicates that the domain is \textit{novel}, not present in the training data of any of the LLMs experimented on.}
\label{tab:domains}
\end{table}

\end{document}

%% file: math_commands.tex
\usepackage{amsmath,amsfonts,bm}

\def\eqref#1{equation~\ref{#1}}

\def\1{\bm{1}}

\DeclareMathAlphabet{\mathsfit}{\encodingdefault}{\sfdefault}{m}{sl}
\SetMathAlphabet{\mathsfit}{bold}{\encodingdefault}{\sfdefault}{bx}{n}

